\documentclass[10pt,journal,cspaper,compsoc]{IEEEtran}

\usepackage{amsmath,amssymb}
\usepackage{algorithm}
\usepackage{algorithmic}
\usepackage{multirow}

\usepackage[aboveskip=8pt]{caption}

\usepackage[dvips]{graphicx}
\DeclareGraphicsExtensions{.pdf}
\usepackage[american]{babel}

\usepackage{tabularx}

\usepackage[bookmarks=false,colorlinks=true,linkcolor=black,citecolor=black,filecolor=black,urlcolor=black]{hyperref}
\usepackage{url}
\usepackage{cvpr}
\usepackage{multicol}

\usepackage{stfloats}

\newcommand{\ve}[1]{\mathbf{#1}} 
\newcommand{\ma}[1]{\mathrm{#1}} 
\newcommand{\tr}{{^{\top}}}
\newcommand{\fr}{_\mathrm{F}}
\newcommand{\tabincell}[2]{\begin{tabular}{@{}#1@{}}#2\end{tabular}}

\newcolumntype{x}{>\small c}
\newcolumntype{y}{>\footnotesize c}
\renewcommand\arraystretch{1.2}

\begin{document}

\title{Accelerating Very Deep Convolutional Networks for Classification and Detection}

\author{Xiangyu~Zhang,
        Jianhua~Zou,
        Kaiming~He$^\dag$\thanks{\dag\quad Correspondence author.},
        and~Jian~Sun
\IEEEcompsocitemizethanks{
\IEEEcompsocthanksitem X. Zhang and J. Zou are with Xi'an Jiaotong University, Xi'an, China. This work was done when X. Zhang was an intern at Microsoft Research.
\IEEEcompsocthanksitem K.~He and J.~Sun are with Microsoft Research, Beijing, China. E-mail: \{kahe,jiansun\}@microsoft.com}
}

\markboth{}%
{Zhang \MakeLowercase{\textit{et al.}}}

\IEEEcompsoctitleabstractindextext{%
\begin{abstract}
This paper aims to accelerate the test-time computation of convolutional neural networks (CNNs), especially very deep CNNs \cite{Simonyan2015} that have substantially impacted the computer vision community.
Unlike previous methods that are designed for approximating linear filters or linear responses, our method takes the nonlinear units into account.
We develop an effective solution to the resulting nonlinear optimization problem without the need of stochastic gradient descent (SGD). More importantly, while previous methods mainly focus on optimizing one or two layers, our nonlinear method enables an asymmetric reconstruction that reduces the rapidly accumulated error when multiple (\eg, $\geq$10) layers are approximated.
For the widely used very deep VGG-16 model \cite{Simonyan2015}, our method achieves a whole-model speedup of 4$\times$ with merely a 0.3\% increase of top-5 error in ImageNet classification. Our 4$\times$ accelerated VGG-16 model also shows a graceful accuracy degradation for object detection when plugged into the Fast R-CNN detector \cite{Girshick2015}.
\end{abstract}

}

\maketitle

\section{Introduction}

The accuracy of convolutional neural networks (CNNs) \cite{LeCun1989,Krizhevsky2012} has been continuously improving \cite{Zeiler2014,Sermanet2014,He2014,Simonyan2015,Szegedy2015}, but the computational cost of these networks also increases significantly. For example, the very deep VGG models \cite{Simonyan2015}, which have witnessed great success in a wide range of recognition tasks \cite{Girshick2014,Girshick2015,Ren2015,Ren2015b,Long2015,Dai2015,Hariharan2015}, are substantially slower than earlier models \cite{Krizhevsky2012,Zeiler2014}.
Real-world systems may suffer from the low speed of these networks.
For example, a cloud service needs to process thousands of new requests per seconds; portable devices such as phones and tablets may not afford slow models; some recognition tasks like object detection \cite{He2014,Girshick2015,Ren2015,Ren2015b} and semantic segmentation \cite{Long2015,Dai2015,Hariharan2015} need to apply these models on higher-resolution images. It is thus of practical importance to accelerate test-time performance of CNNs.

There have been a series of studies on accelerating deep CNNs \cite{Vanhoucke2011,Denton2014,Jaderberg2014,Lebedev2015}. A common focus of these methods is on the decomposition of one or a few layers. These methods have shown promising speedup ratios and accuracy on one or two layers and whole (but shallower) models. However, few results are available for accelerating \emph{very deep} models (\eg, $\geq$ 10 layers). Experiments on complex datasets such as ImageNet \cite{Russakovsky2014} are also limited - \eg, the results in \cite{Denton2014,Jaderberg2014,Lebedev2015} are about accelerating a \emph{single} layer of the shallower AlexNet \cite{Krizhevsky2012}. Moreover, performance of the accelerated networks as generic feature extractors for other recognition tasks \cite{Girshick2015,Long2015} remain unclear.

It is nontrivial to speed up \emph{whole}, \emph{very deep} models for \emph{complex} tasks like ImageNet classification. Acceleration algorithms involve not only the decomposition of layers, but also the optimization solutions to the decomposition. Data (response) reconstruction solvers \cite{Jaderberg2014} based on stochastic gradient descent (SGD) and backpropagation work well for simpler tasks such as character classification \cite{Jaderberg2014}, but are less effective for complex ImageNet models (as we will discussed in Sec.~\ref{sec:exp}). These SGD-based solvers are sensitive to initialization and learning rates, and might be trapped into poorer local optima for regressing responses.
Moreover, even when a solver manages to accelerate a single layer, the \emph{accumulated} error of approximating multiple layers grow rapidly, especially for very deep models. Besides, the layers of a very deep model may exhibit a great diversity in filter numbers, feature map sizes, sparsity, and redundancy. It may not be beneficial to uniformly accelerate all layers.

In this paper, we present an accelerating method that is effective for very deep models. We first propose a response reconstruction method that takes into account the nonlinear neurons and a low-rank constraint. A solution based on Generalized Singular Value Decomposition (GSVD) is developed for this nonlinear problem, without the need of SGD.
Our explicit treatment of the nonlinearity better models a nonlinear layer, and more importantly, enables an \emph{asymmetric} reconstruction that accounts for the error from previous approximated layers. This method effectively reduces the accumulated error when multiple layers are approximated sequentially. We also present a rank selection method for adaptively determining the acceleration of each layer for a whole model, based on their redundancy.

In experiments, we demonstrate the effects of the nonlinear solution, asymmetric reconstruction, and whole-model acceleration by controlled experiments of a 10-layer model on ImageNet classification \cite{Russakovsky2014}.
Furthermore, we apply our method on the publicly available \textbf{VGG-16} model \cite{Simonyan2015}, and achieve a 4$\times$ speedup with merely a \textbf{0.3\%} increase of top-5 center-view error.

The impact of the ImageNet dataset \cite{Russakovsky2014} is not merely on the specific 1000-class classification task; deep models pre-trained on ImageNet have been actively used to replace hand-engineered features, and have showcased excellent accuracy for challenging tasks such as object detection \cite{Girshick2014,Girshick2015,Ren2015,Ren2015b} and semantic segmentation \cite{Long2015,Dai2015,Hariharan2015}. We exploit our method to accelerate the very deep VGG-16 model for Fast R-CNN \cite{Girshick2015} object detection. With a 4$\times$ speedup of all convolutions, our method has a graceful degradation of 0.8\% mAP (from 66.9\% to 66.1\%) on the PASCAL VOC 2007 detection benchmark \cite{Everingham2007}.

A preliminary version of this manuscript has been presented in a conference \cite{Zhang2015}. This manuscript extends the initial version from several aspects to strengthen our method. (1) We demonstrate compelling acceleration results on very deep VGG models, and are among the first few works accelerating very deep models. (2) We investigate the accelerated models for transfer-learning-based object detection \cite{Girshick2014,Girshick2015}, which is one of the most important applications of ImageNet pre-trained networks. (3) We provide evidence showing that a model trained from scratch and sharing the same structure as the accelerated model is inferior. This discovery suggests that a very deep model can be accelerated not simply because the decomposed network architecture is more powerful, but because the acceleration optimization algorithm is able to digest information.

\section{Related Work}

Methods \cite{Vanhoucke2011,Denton2014,Jaderberg2014,Lebedev2015} for accelerating test-time computation of CNNs in general have two components: (i) a layer decomposition design that reduces time complexity, and (ii) an optimization scheme for the decomposition design. Although the former (``decomposition'') attracts more attention because it directly addresses the time complexity, the latter (``optimization'') is also essential because not all decompositions are similarly easy to fine good local optima.

The method of Denton \etal \cite{Denton2014} is one of the first to exploit low-rank decompositions of filters. Several decomposition designs along different dimensions have been investigated.
This method does not explicitly minimize the error of the activations after the nonlinearity, which is influential to the accuracy as we will show. This method presents experiments of accelerating a single layer of an OverFeat network \cite{Sermanet2014}, but no whole-model results are available.

Jaderberg \etal \cite{Jaderberg2014} present efficient decompositions by separating $k \times k$ filters into $k \times 1$ and $1 \times k$ filters, which was earlier developed for accelerating generic image filters \cite{Rigamonti2013}. Channel-wise dimension reduction is also considered. Two optimization schemes are proposed: (i) ``filter reconstruction'' that minimizes the error of filter weights, and (ii) ``data reconstruction'' that minimizes the error of responses. In \cite{Jaderberg2014}, conjugate gradient descent is used to solve filter reconstruction, and SGD with backpropagation is used to solve data reconstruction. Data reconstruction in \cite{Jaderberg2014} demonstrates excellent performance on a character classification task using a 4-layer network. For ImageNet classification, their paper evaluates a single layer of an OverFeat network by ``filter reconstruction''. But the performance of whole, very deep models in ImageNet remains unclear.

Concurrent with our work, Lebedev \etal \cite{Lebedev2015} adopt ``CP-decomposition'' to decompose a layer into five layers of lower complexity. For ImageNet classification, only a single-layer acceleration of AlexNet is reported in \cite{Lebedev2015}.
Moreover, Lebedev \etal report that they ``failed to find a good SGD learning rate'' in their fine-tuning, suggesting that it is nontrivial to optimize the factorization for even a single layer in ImageNet models.

Despite some promising preliminary results that have been obtained in the above works \cite{Denton2014,Jaderberg2014,Lebedev2015}, the \emph{whole-model} acceleration of \emph{very deep} networks for \emph{ImageNet} is still an open problem.

Besides the research on decomposing layers, there have been other streams on improving train/test-time performance of CNNs. FFT-based algorithms \cite{Vasilache2015,Mathieu2013} are applicable for both training and testing, and are particularly effective for large spatial kernels. On the other hand, it is also proposed to train ``thin'' and deep networks \cite{He2015a,Romero2015} for good trade-off between speed and accuracy. Besides reducing running time, a related issue involving memory conservation \cite{Collins2014} has also attracted attention.

\section{Approaches}
\label{sec:method}

Our method exploits a low-rank assumption for decomposition, following the stream of \cite{Denton2014,Jaderberg2014}. We show that this decomposition has a closed-form solution (SVD) for linear neurons, and a slightly more complicated solution (GSVD \cite{Gower2004,Takane2006,Takane2007}) for nonlinear neurons. The simplicity of our solver enables an asymmetric reconstruction method for reducing accumulated error of very deep models.

\begin{figure}[t]
\begin{center}
\includegraphics[width=0.85\linewidth]{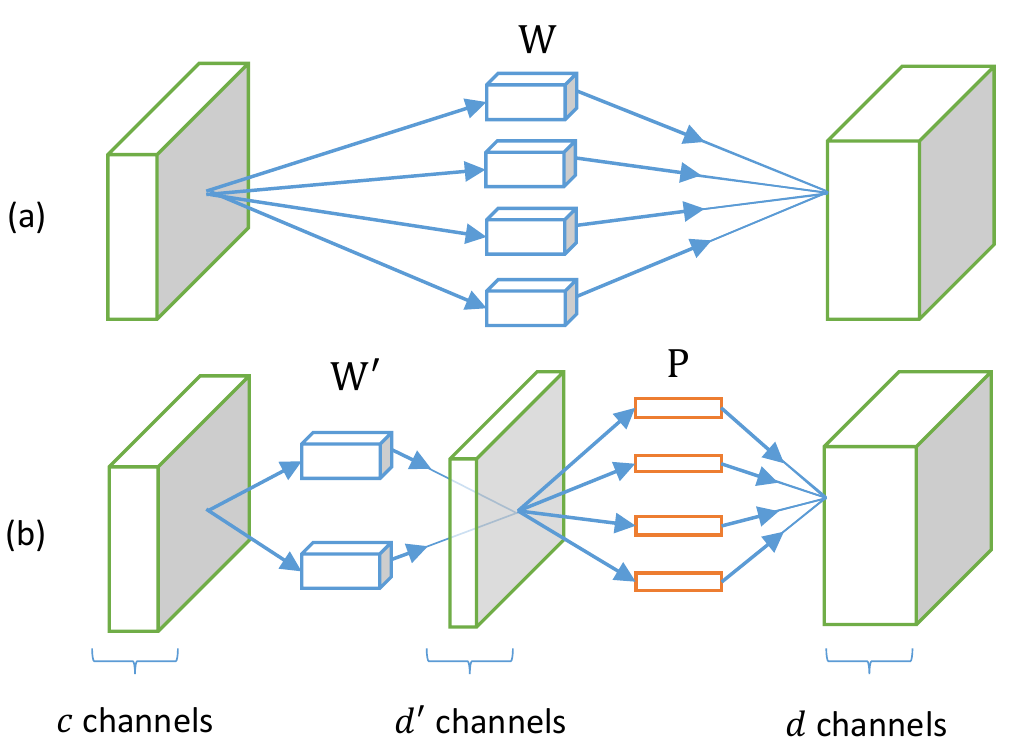}
\end{center}
   \caption{Illustration of the decomposition. (a) An original layer with complexity $O(dk^2c)$. (b) An approximated layer with complexity reduced to $O(d'k^2c)+O(dd')$.}
\label{fig:concept}
\end{figure}

\subsection{Low-rank Approximation of Responses}
\label{sec:linear}

Our assumption is that the filter response at a pixel of a layer approximately lies on a low-rank subspace. A resulting low-rank decomposition reduces time complexity.
To find the approximate low-rank subspace, we minimize the reconstruction error of the responses.

More formally, we consider a convolutional layer with a filter size of $k\times k\times c$, where $k$ is the spatial size of the filter and $c$ is the number of input channels of this layer. To compute a response, this filter is applied on a $k\times k\times c$ volume of the layer input. We use $\ve{x}\in\mathbb{R}^{k^2c+1}$ to denote a vector that reshapes this volume, where we append one as the last entry for the sake of the bias. A response $\ve{y}\in\mathbb{R}^{d}$ at a position of a layer is computed as:
\begin{equation}\label{eq:y}
\ve{y}=\ma{W}\ve{x}.
\end{equation}
where $\ma{W}$ is a $d$-by-($k^2c$$+$$1$) matrix, and $d$ is the number of filters.
Each row of $\ma{W}$ denotes the reshaped form of a $k\times k\times c$ filter with the bias appended. 

Under the assumption that the vector $\ve{y}$ is on a low-rank subspace, we can write $\ve{y}=\ma{M}(\ve{y}-\bar{\ve{y}})+\bar{\ve{y}}$, where $\ma{M}$ is a $d$-by-$d$ matrix of a rank $d'<d$ and $\bar{\ve{y}}$ is the mean vector of responses. Expanding this equation, we can compute a response by:
\begin{equation}\label{eq:y1}
\ve{y}=\ma{M}\ma{W}\ve{x}+\ve{b},
\end{equation}
where $\ve{b}=\bar{\ve{y}}-\ma{M}\bar{\ve{y}}$ is a new bias. The rank-$d'$ matrix $\ma{M}$ can be decomposed into two $d$-by-$d'$ matrices $\ma{P}$ and $\ma{Q}$ such that $\ma{M}=\ma{P}\ma{Q}\tr$.
We denote $\ma{W}'=\ma{Q}\tr\ma{W}$ as a $d'$-by-($k^2c$$+$$1$) matrix, which is essentially a new set of $d'$ filters. Then we can compute (\ref{eq:y1}) by:
\begin{equation}\label{eq:y2}
\ve{y}=\ma{P}\ma{W}'\ve{x}+\ve{b}.
\end{equation}
The complexity of using Eqn.(\ref{eq:y2}) is $O(d'k^2c)+O(dd')$, while the complexity of using Eqn.(\ref{eq:y}) is $O(dk^2c)$. For many typical models/layers, we usually have $O(dd')\ll O(d'k^2c)$, so the computation in Eqn.(\ref{eq:y2}) will reduce the complexity to about $d'/d$.

Fig.~\ref{fig:concept} illustrates how to use Eqn.(\ref{eq:y2}) in a network. We replace the original layer (given by $\ma{W}$) by two layers (given by $\ma{W}'$ and $\ma{P}$). The matrix $\ma{W}'$ is actually $d'$ filters whose sizes are $k\times k\times c$. These filters produce a $d'$-dimensional feature map. On this feature map, the $d$-by-$d'$ matrix $\ma{P}$ can be implemented as $d$ filters whose sizes are $1\times 1\times d'$. So $\ma{P}$ corresponds to a convolutional layer with a 1$\times$1 spatial support, which maps the $d'$-dimensional feature map to a $d$-dimensional one.

\begin{figure*}
\begin{center}
\includegraphics[width=0.98\linewidth]{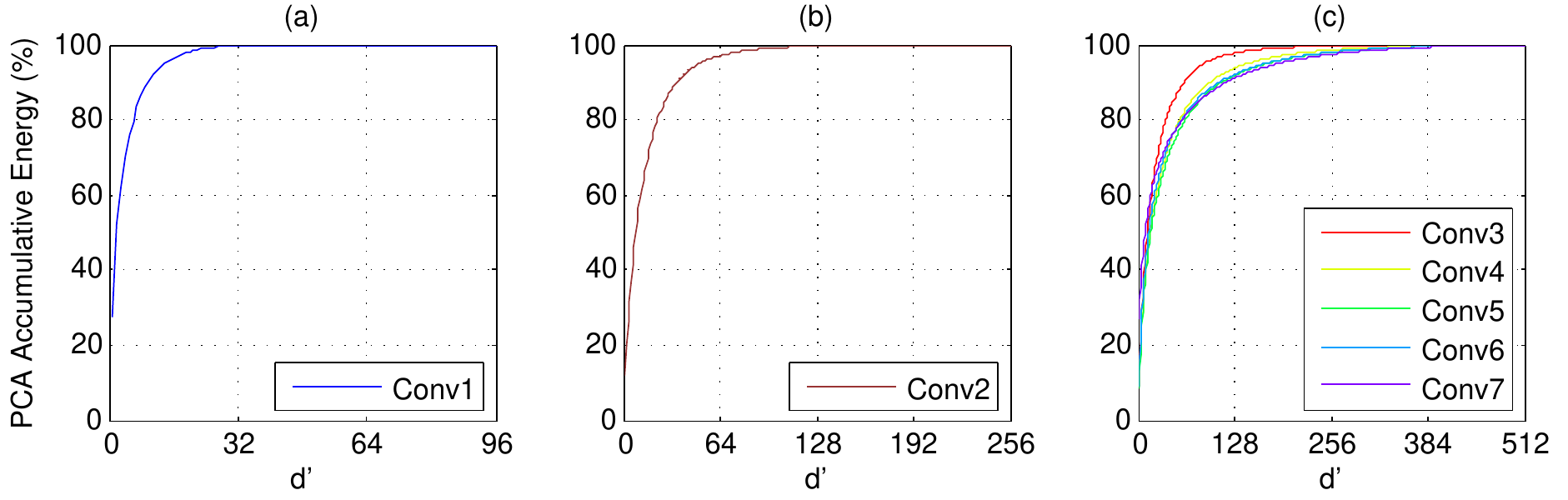}
\end{center}
   \caption{PCA accumulative energy of the responses in each layer, presented as the sum of largest $d'$ eigenvalues (relative to the total energy when $d'=d$). Here the filter number $d$ is 96 for Conv1, 256 for Conv2, and 512 for Conv3-7 (detailed in Table~\ref{tbl:arch}). These figures are obtained from 3,000 randomly sampled training images.}
\label{fig:pca_energy}
\end{figure*}

Note that the decomposition of $\ma{M}=\ma{P}\ma{Q}\tr$ can be arbitrary. It does not impact the value of $\ve{y}$ computed in Eqn.(\ref{eq:y2}). A simple decomposition is the Singular Value Decomposition (SVD) \cite{Golub1996}: $\ma{M}=\ma{U}_{d'}\ma{S}_{d'}\ma{V}_{d'}\tr$, where $\ma{U}_{d'}$ and $\ma{V}_{d'}$ are $d$-by-$d'$ column-orthogonal matrices and $\ma{S}_{d'}$ is a $d'$-by-$d'$ diagonal matrix. Then we can obtain $\ma{P}=\ma{U}_{d'}\ma{S}^{{1}/{2}}_{d'}$ and $\ma{Q}=\ma{V}_{d'}\ma{S}^{{1}/{2}}_{d'}$.

In practice the low-rank assumption does not strictly hold, and the computation in Eqn.(\ref{eq:y2}) is approximate.
To find an approximate low-rank subspace, we optimize the following problem:
\begin{gather}\label{eq:pca}
\min_{\ma{M}}\sum_i\|(\ve{y}_i-\bar{\ve{y}})-\ma{M}(\ve{y}_i-\bar{\ve{y}})\|^2_2,\\
s.t.\quad rank(\ma{M})\leq d'.\nonumber
\end{gather}
Here $\ve{y}_i$ is a response sampled from the feature maps in the training set.
This problem can be solved by SVD \cite{Golub1996} or actually Principal Component Analysis (PCA): let $\ma{Y}$ be the $d$-by-$n$ matrix concatenating $n$ responses with the mean subtracted, compute the eigen-decomposition of the covariance matrix $\ma{Y}\ma{Y}\tr=\ma{U}\ma{S}\ma{U}\tr$ where $\ma{U}$ is an orthogonal matrix and $\ma{S}$ is diagonal, and $\ma{M}=\ma{U}_{d'}\ma{U}_{d'}\tr$ where $\ma{U}_{d'}$ are the first $d'$ eigenvectors. With the matrix $\ma{M}$ computed, we can find $\ma{P}=\ma{Q}=\ma{U}_{d'}$.

How good is the low-rank assumption? We sample the responses from a CNN model (with 7 convolutional layers, detailed in Sec.~\ref{sec:exp}) trained on ImageNet. For the responses of each layer, we compute the eigenvalues of their covariance matrix and then plot the sum of the largest eigenvalues (Fig.~\ref{fig:pca_energy}). We see that substantial energy is in a small portion of the largest eigenvectors. For example, in the Conv2 layer ($d=256$) the first 128 eigenvectors contribute over 99.9\% energy; in the Conv7 layer ($d=512$), the first 256 eigenvectors contribute over 95\% energy.
This indicates that we can use a fraction of the filters to precisely approximate the original filters.

The low-rank behavior of the responses $\ve{y}$ is because of the low-rank behaviors of the filter weights $\ma{W}$ and the inputs $\ve{x}$. Although the low-rank assumptions about filter weights $\ma{W}$ have been adopted in recent work \cite{Denton2014,Jaderberg2014}, we further adopt the low-rank assumptions about the filter inputs $\ve{x}$, which are local volumes and have correlations. The responses $\ve{y}$ will have lower rank than $\ma{W}$ and $\ve{x}$, so the approximation can be more precise.
In our optimization (\ref{eq:pca}), we directly address the low-rank subspace of $\ve{y}$.

\subsection{Nonlinear Case}
\label{sec:nonlinear}

Next we investigate the case of using nonlinear units. We use $r(\cdot)$ to denote the nonlinear operator. In this paper we focus on the Rectified Linear Unit (ReLU) \cite{Nair2010}: $r(\cdot)=\max(\cdot, 0)$. 

Driven by Eqn.(\ref{eq:pca}), we minimize the reconstruction error of the nonlinear responses:
\begin{gather}\label{eq:relu}
\min_{\ma{M},\ve{b}}\sum_i\|r(\ve{y}_i)-r(\ma{M}\ve{y}_i+\ve{b})\|^2_2,\\
s.t.\quad rank(\ma{M})\leq d'.\nonumber
\end{gather}
Here $\ve{b}$ is a new bias to be optimized, and $r(\ma{M}\ve{y}+\ve{b})=r(\ma{M}\ma{W}\ve{x}+\ve{b})$ is the nonlinear response computed by the approximated filters.

The above optimization problem is challenging due to the nonlinearity and the low-rank constraint.
To find a feasible solution, we relax it as:
\begin{gather}
\min_{\ma{M},\ve{b},\{\ve{z}_i\}}\sum_i\|r(\ve{y}_i)-r(\ve{z}_i)\|^2_2
+\lambda\|\ve{z}_i-(\ma{M}\ve{y}_i+\ve{b})\|^2_2\nonumber\\
s.t.\quad rank(\ma{M})\leq d'.
\label{eq:relu1}
\end{gather}
Here $\{\ve{z}_i\}$ is a set of auxiliary variables of the same size as $\{\ve{y}_i\}$. $\lambda$ is a penalty parameter. If $\lambda\rightarrow \infty$, the solution to (\ref{eq:relu1}) will converge to the solution to (\ref{eq:relu}) \cite{Wang2008}. We adopt an alternating solver, fixing $\{\ve{z}_i\}$ and solving for $\ma{M}$, $\ve{b}$ and vice versa.

\vspace{8pt}
\noindent \textbf{(i) The subproblem of $\ma{M}$, $\ve{b}$}. In this case, $\{\ve{z}_i\}$ are fixed.
It is easy to show that $\ve{b}$ is solved by $\ve{b}=\bar{\ve{z}}-\ma{M}\bar{\ve{y}}$ where $\bar{\ve{z}}$ is the mean vector of $\{\ve{z}_i\}$. Substituting $\ve{b}$ into the objective function, we obtain the problem involving $\ma{M}$:
\begin{gather}\label{eq:pcaz}
\min_{\ma{M}}\sum_i\|(\ve{z}_i-\bar{\ve{z}})-\ma{M}(\ve{y}_i-\bar{\ve{y}})\|^2_2,\\
s.t.\quad rank(\ma{M})\leq d'.\nonumber
\end{gather}
This problem appears similar to Eqn.(\ref{eq:pca}) except that there are two sets of responses.

This optimization problem also has a closed-form solution by Generalized SVD (GSVD) \cite{Gower2004,Takane2006,Takane2007}.
Let $\ma{Z}$ be the $d$-by-$n$ matrix concatenating the vectors of $\{\ve{z}_i-\bar{\ve{z}}\}$. We rewrite the above problem as:
\begin{gather}\label{eq:rank}
\min_{\ma{M}}\|\ma{Z}-\ma{M}\ma{Y}\|^2\fr,\\
s.t.\quad rank(\ma{M})\leq d'.\nonumber
\end{gather}
Here $\|\cdot\|\fr$ is the Frobenius norm.
A problem in this form is known as Reduced Rank Regression \cite{Gower2004,Takane2006,Takane2007}.
This problem belongs to a broader category of \emph{procrustes} problems \cite{Gower2004} that have been adopted for various data reconstruction problems \cite{Gong2011,Ge2014,Xia2015}.
The solution is as follows (see \cite{Takane2007}). Let $\ma{\hat{M}}=\ma{Z}\ma{Y}\tr(\ma{Y}\ma{Y}\tr)^{-1}$. GSVD \cite{Takane2007} is applied on $\ma{\hat{M}}$: $\ma{\hat{M}}=\ma{U}\ma{S}\ma{V}\tr$, such that $\ma{U}$ is a $d$-by-$d$ orthogonal matrix satisfying $\ma{U}\tr\ma{U}=\ma{I}_d$ where $\ma{I}_d$ is a $d$-by-$d$ identity matrix, and
$\ma{V}$ is a $d$-by-$d$ matrix satisfying $\ma{V}\tr \ma{Y}\ma{Y}\tr \ma{V=\ma{I}_d}$ (called \emph{generalized orthogonality}).
Then the solution $\ma{M}$ to (\ref{eq:rank}) is given by $\ma{M}=\ma{U}_{d'}\ma{S}_{d'}\ma{V}_{d'}\tr$ where $\ma{U}_{d'}$ and $\ma{V}_{d'}$ are the first $d'$ columns of $\ma{U}$ and $\ma{V}$ and $\ma{S}_{d'}$ are the largest $d'$ singular values. One can show that if $\ma{Z}=\ma{Y}$ (so the problem in (\ref{eq:pcaz}) becomes (\ref{eq:pca})), this GSVD solution becomes SVD, \ie, eigen-decomposition of $\ma{Y}\ma{Y}\tr$.

\vspace{8pt}
\noindent \textbf{(ii) The subproblem of $\{\ve{z}_i\}$}.
In this case, $\ma{M}$ and $\ve{b}$ are fixed. Then in this subproblem each element $z_{ij}$ of each vector $\ve{z}_i$ is independent of any other. So we solve a 1-dimensional optimization problem as follows:
\begin{gather}
\min_{z_{ij}}~(r(y_{ij})-r(z_{ij}))^2+\lambda(z_{ij}-y'_{ij})^2,
\label{eq:relu2}
\end{gather}
where $y'_{ij}$ is the $j$-th entry of $\ma{M}\ve{y}_i+\ve{b}$. By separately considering $z_{ij}\geq0$ and $z_{ij}<0$, we obtain the solution as follows: let
\begin{gather}
z_{0} = \min (0, y'_{ij})\\
z_{1} = \max (0, \frac{\lambda\cdot y'_{ij} + r(y_{ij})}{\lambda + 1})
\end{gather}
then $z_{ij}=\arg\min_{z_{0},z_{1}}(r(y_{ij})-r(z_{ij}))^2+\lambda(z_{ij}-y'_{ij})^2$.
Our method is also applicable for other types of nonlinearities. The subproblem in (\ref{eq:relu2}) is a 1-dimensional nonlinear least squares problem, so can be solved by gradient descent for other $r(\cdot)$.

\vspace{12pt}
We alternatively solve (i) and (ii).
The initialization is given by the solution to the linear case (\ref{eq:pca}). We warm up the solver by setting the penalty parameter $\lambda=0.01$ and run 25 iterations. Then we increase the value of $\lambda$. In theory,
$\lambda$ should be gradually increased to infinity \cite{Wang2008}. But
we find that it is difficult for the iterative solver to make progress if $\lambda$ is too large. So we increase $\lambda$ to 1, run 25 more iterations, and use the resulting $\ma{M}$ as our solution.
As before, we obtain $\ma{P}$ and $\ma{Q}$ by SVD on $\ma{M}$.

In experiments, we find that it is sufficient to randomly sample 3,000 images to solve Eqn.(\ref{eq:relu}). It only takes our method \textbf{2-5 minutes} in MATLAB solving a layer. This is much faster than SGD-based solvers.

\subsection{Asymmetric Reconstruction for Multi-Layer}
\label{sec:asymmetric}

When each layer is approximated independently, the error of shallower layers will be rapidly accumulated and affect deeper layers. We propose an asymmetric reconstruction method to alleviate this problem.

We apply our method sequentially on each layer, from the shallower layers to the deeper ones.
Let us consider a layer whose input feature map is not precise due to the approximation of the previous layer/layers. We denote the approximate input to the current layer as $\ve{\hat{x}}$. For the training data, we can still compute its non-approximate responses as $\ve{y}=\ma{W}\ve{x}$. So we can optimize an ``asymmetric'' version of (\ref{eq:relu}):
\begin{gather}\label{eq:reluasy}
\min_{\ma{M},\ve{b}}\sum_i\|r(\ma{W}\ve{x}_i)-r(\ma{M}\ma{W}\ve{\hat{x}}_i+\ve{b})\|^2_2,\\
s.t.\quad rank(\ma{M})\leq d'.\nonumber
\end{gather}
In the first term $r(\ma{W}\ve{x})=r(\ve{y})$ is the non-approximate output of this layer. In the second term, $\ve{\hat{x}}_i$ is the approximated input to this layer, and $r(\ma{M}\ma{W}\ve{\hat{x}}_i+\ve{b})$ is the approximated output of this layer. In contrast to using $\ve{x}$ (or $\ve{\hat{x}}$) for both terms, this asymmetric formulation faithfully incorporates the two actual terms before/after the approximation of this layer.
The optimization problem in (\ref{eq:reluasy}) can be solved using the same algorithm as for (\ref{eq:relu}).

\renewcommand{\arraystretch}{1.3}
\begin{table*}
\begin{center}
\small
\begin{tabular}{|c|c|c|c|c|c|c|c|c|}
\hline
layer & filter size & \# channels & \# filters & stride & output size & complexity (\%) & \# of zeros \\
\hline
Conv1 & 7 $\times$ 7 & 3 & 96 & 2 & 109 $\times$ 109 & 3.8 & 0.49 \\
Pool1 & 3 $\times$ 3 &   &    & 3 & 37 $\times$ 37   &     &      \\
\hline
Conv2 & 5 $\times$ 5 & 96 & 256 & 1 & 35 $\times$ 35 & 17.3 & 0.62 \\
Pool2 & 2 $\times$ 2 &   &    & 2 & 18 $\times$ 18   &     &      \\
\hline
Conv3 & 3 $\times$ 3 & 256 & 512 & 1 & 18 $\times$ 18 & 8.8 & 0.60 \\
Conv4 & 3 $\times$ 3 & 512 & 512 & 1 & 18 $\times$ 18 & 17.5 & 0.69 \\
Conv5 & 3 $\times$ 3 & 512 & 512 & 1 & 18 $\times$ 18 & 17.5 & 0.69 \\
Conv6 & 3 $\times$ 3 & 512 & 512 & 1 & 18 $\times$ 18 & 17.5 & 0.68 \\
Conv7 & 3 $\times$ 3 & 512 & 512 & 1 & 18 $\times$ 18 & 17.5 & 0.95 \\
\hline
\end{tabular}
\end{center}
\caption{The architecture of the SPP-10 model \cite{He2014}. It has 7 conv layers and 3 fc layers. Each layer (except the last fc) is followed by ReLU. The final conv layer is followed by a spatial pyramid pooling layer \cite{He2014} that have 4 levels ($\{6\times6, 3\times3, 2\times2, 1\times1\}$, totally 50 bins). The resulting $50\times512$-d is fed into the 4096-d fc layer (fc6), followed by another 4096-d fc layer (fc7) and a 1000-way softmax layer. The column ``complexity'' is the theoretical time complexity, shown as relative numbers to the total convolutional complexity. The column ``\# of zeros'' is the relative portion of zero responses, which shows the ``sparsity'' of the layer.}
\label{tbl:arch}
\end{table*}

\begin{figure}
\begin{center}
\includegraphics[width=0.98\linewidth]{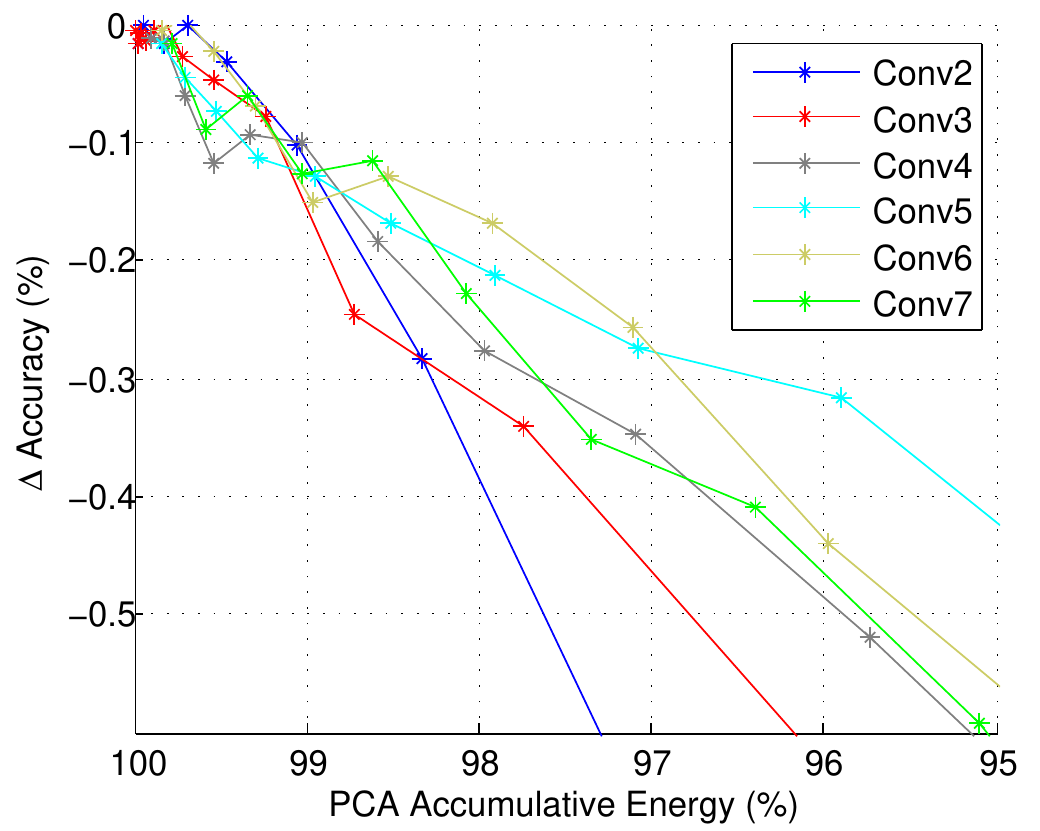}
\end{center}
   \caption{PCA accumulative energy and the accuracy rates (top-5). Here the accuracy is evaluated using the linear solution (the nonlinear solution has a similar trend). Each layer is evaluated independently, with other layers not approximated. The accuracy is shown as the difference to no approximation.}
\label{fig:pca_sigma}
\end{figure}

\subsection{Rank Selection for Whole-Model Acceleration}
\label{sec:parameterselection}

In the above, the optimization is based on a target $d'$ of each layer. $d'$ is the only parameter that determines the complexity of an accelerated layer.
But given a desired speedup ratio of the \emph{whole model}, we need to determine the proper rank $d'$ used for each layer. One may adopt a uniform speedup ratio for each layer. But this is not an optimal solution, because the layers are not equally redundant.

We empirically observe that the PCA energy after approximations is roughly related to the classification accuracy.
To verify this observation, in Fig.~\ref{fig:pca_sigma} we show the classification accuracy (represented as the difference to no approximation) \vs the PCA energy.
Each point in this figure is empirically evaluated using a reduced rank $d'$. 100\% energy means no approximation and thus no degradation of classification accuracy.
Fig.~\ref{fig:pca_sigma} shows that the classification accuracy is roughly linear on the PCA energy.

To simultaneously determine the reduced ranks of all layers, we further assume that the whole-model classification accuracy is roughly related to the product of the PCA energy of all layers. More formally, we consider this objective function:
\begin{gather}
\mathcal{E}=\prod_{l}\sum_{a=1}^{d'_{l}}{\sigma_{l,a}}
\end{gather}
Here $\sigma_{l,a}$ is the $a$-th largest eigenvalue of the layer $l$, and $\sum_{a=1}^{d'_{l}}{\sigma_{l,a}}$ is the PCA energy of the largest $d'_{l}$ eigenvalues in the layer $l$. The product $\prod_{l}$ is over all layers to be approximated. The objective $\mathcal{E}$ is assumed to be related to the accuracy of the approximated whole network.
Then we optimize this problem:
\begin{gather}\label{eq:compcost}
\max_{\{d'_l\}}\mathcal{E},\quad\quad
s.t.\quad \sum_l{\frac{d'_l}{d_l}C_l} \leq C.
\end{gather}
Here $d_l$ is the original number of filters in the layer $l$, and $C_l$ is the original time complexity of the layer $l$. So $\frac{d'_l}{d_l}C_l$ is the complexity after the approximation. $C$ is the total complexity after the approximation, which is given by the desired speedup ratio. This optimization problem means that we want to maximize the accumulated energy subject to the time complexity constraint.

The problem in (\ref{eq:compcost}) is a combinatorial problem \cite{Reeves1993}.
So we adopt a greedy strategy to solve it. We initialize $d'_l$ as $d_l$, and consider the set $\{\sigma_{l,a}\}$. In each step we remove an eigenvalue $\sigma_{l,d'_l}$ from this set, chosen from a certain layer $l$. The relative reduction of the objective is $\triangle \mathcal{E} / \mathcal{E}=\sigma_{l,d'}/{\sum_{a=1}^{d'_l}\sigma_{l,a}}$, and the reduction of complexity is $\triangle C = {\frac{1}{d_l}C_l}$. Then we define a measure as $\frac{\triangle \mathcal{E} / \mathcal{E}}{\triangle C}$.
The eigenvalue $\sigma_{l,d'_l}$ that has the smallest value of this measure is removed. Intuitively, this measure favors a small reduction of $\triangle \mathcal{E} / \mathcal{E}$ and a large reduction of complexity $\triangle C$. This step is greedily iterated, until the constraint of the total complexity is achieved.

\subsection{Higher-Dimensional Decomposition}
\label{sec:hd}

In our formulation, we focus on reducing the channels (from $d$ to $d'$).
There are algorithmic advantages of operating on the channel dimension. Firstly, this dimension can be easily controlled by the rank constraint $rank(\ma{M})\leq d'$. This constraint enables closed-form solutions, \eg, SVD or GSVD.
Secondly, the optimized low-rank projection $\ma{M}$ can be exactly decomposed into low-dimensional filters ($\ma{P}$ and $\ma{Q}$). These simple and closed-form solutions can produce good results using a very small subset of training images (3,000 out of one million).

On the other hand, compared with decomposition methods that operate on multiple dimensions (spatial and channel) \cite{Jaderberg2014}, our method has to use a smaller $d'$ to approach a given speedup ratio, which might limit the accuracy of our method. To avoid $d'$ being too small, we further propose to combine our solver with Jaderberg \etal's spatial decomposition. Thanks to our asymmetric reconstruction, our method can effectively alleviate the accumulated error for the multi-decomposition.

To determined the decomposed architecture (but not yet the weights), we first use our method to decompose all conv layers of a model. This involves the rank selection of $d'$ for all layers. Then we apply Jaderberg \etal's method to further decompose the resulting $k\times k$ layers ($k > 1$) into $k\times1$ and $1\times k$ filters. The first $k\times1$ layer has $d''$ output channels depending on the speedup ratio. In this way, an original layer of ($k \times k$, $d$) is decomposed into three layers of ($k \times 1$, $d''$), ($1 \times k$, $d'$), and ($1 \times 1$, $d$). For a speedup ratio $r$, we let each method contribute a speedup of $\sqrt{r}$.

With the decomposed architecture determined, we solve for the weights of the decomposed layers. Given their order as above, we first optimize the ($k \times 1$, $d''$) and ($1 \times k$, $d$) layers using ``filter reconstruction'' \cite{Jaderberg2014} (we will discuss ``data reconstruction'' later). Then we adopt our solution on the ($1\times k$, $d$) layer and optimize for the ($1 \times k$, $d'$) and ($1 \times 1$, $d$) layers. We use our asymmetric reconstruction in Eqn.(\ref{eq:reluasy}). In the $r(\ma{M}\ma{W}\ve{\hat{x}}+\ve{b})$ term, $\hat{x}$ is the approximated input to this $1\times k$ layer, and the $r(\ma{W}\ve{x})=r(\ve{y})$ term is still the true response of the original $k \times k$ layer without any decomposition. The approximation error of the spatial decomposition will also be addressed by our asymmetric reconstruction, which is important to alleviate accumulated error.
We term this as ``asymmetric (3d)'' in the following.

\begin{figure*}[t]
\begin{center}
\includegraphics[width=0.98\linewidth]{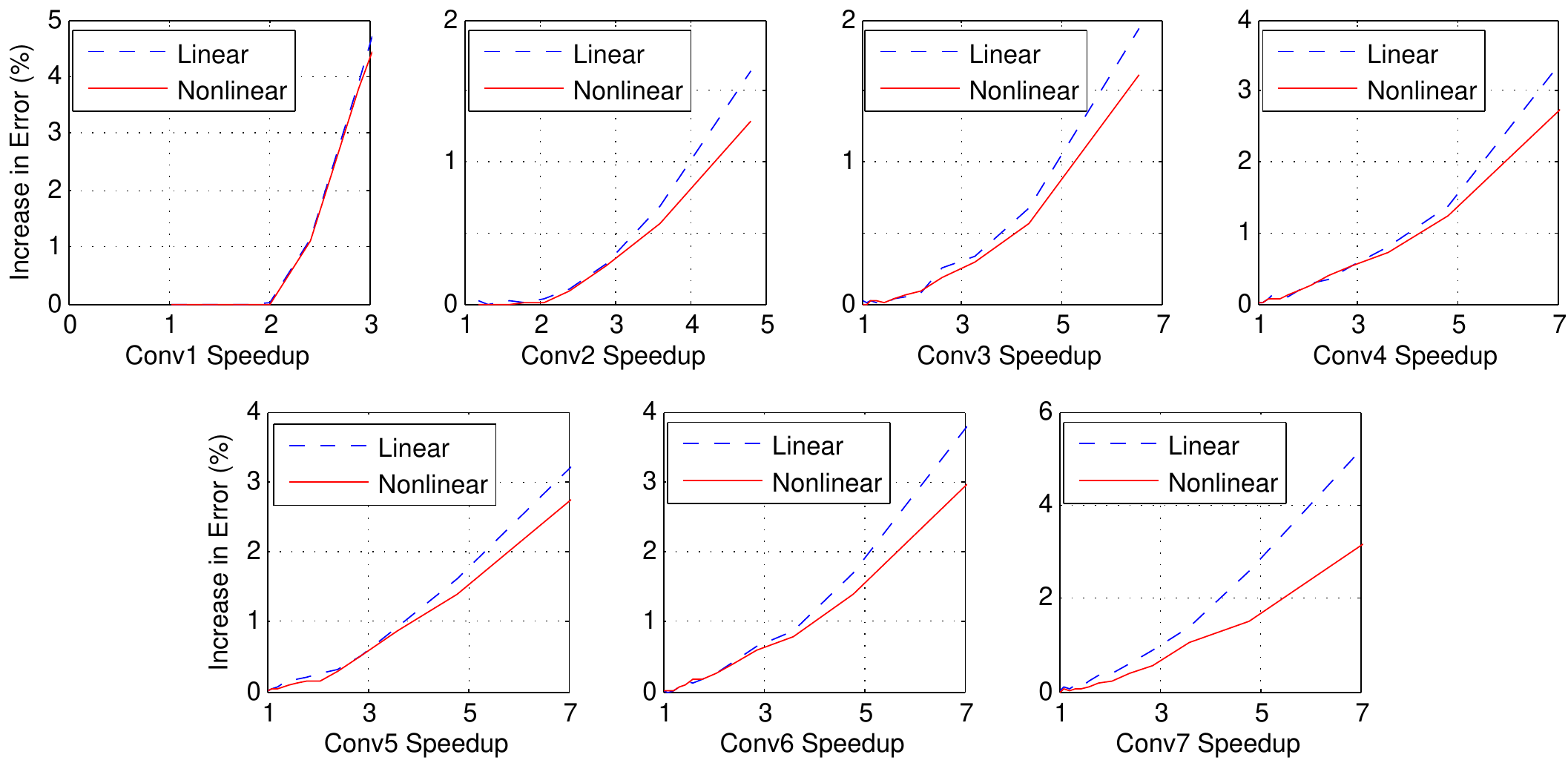}
\end{center}
   \caption{\textbf{Linear vs. Nonlinear} for SPP-10: single-layer performance of accelerating Conv1 to Conv7. The speedup ratios are computed by the theoretical complexity of that layer. The error rates are top-5 single-view, and shown as the increase of error rates compared with no approximation (\emph{smaller is better}).}
\label{fig:layerwiseresult}
\end{figure*}

\subsection{Fine-tuning}
\label{sec:ft}

With any approximated whole model, we may ``fine-tune'' this model end-to-end in the ImageNet training data. This process is similar to training a classification network with the approximated model as the initialization.

However, we empirically find that fine-tuning is very sensitive to the initialization (given by the approximated model) and the learning rate. If the initialization is poor and the learning rate is small, the fine-tuning is easily trapped in a poor local optimum and makes little progress. If the learning rate is large, the fine-tuning process behaves very similar to training the decomposed architecture ``from scratch'' (as we will discuss later). A large learning rate may jump out of the initialized local optimum, and the initialization appears to be ``forgotten''.

Fortunately, our method has achieved very good accuracy even without fine-tuning as we will show by experiments. With our approximated model as the initialization, the fine-tuning with a sufficiently small learning rate is able to further improve the results. In our experiments, we use a learning rate of 1e-5 and a mini-batch size of 128, and fine-tune the models for 5 epochs in the ImageNet training data.

We note that in the following the results are \textbf{without} fine-tuning unless specified.

\section{Experiments}
\label{sec:exp}

We comprehensively evaluate our method on two models. The first model is a 10-layer model of ``SPPnet (OverFeat-7)'' in \cite{He2014}, which we denote as ``\textbf{SPP-10}''.
This model (detailed in Table~\ref{tbl:arch}) has a similar architecture to the OverFeat model \cite{Sermanet2014} but is deeper. It has 7 conv layers and 3 fc layers.
The second model is the publicly available \textbf{VGG-16} model \cite{Simonyan2015}\footnote{\url{www.robots.ox.ac.uk/~vgg/research/very_deep/}} that has 13 conv layers and 3 fc layers.
\textbf{SPP-10} won the 3-rd place and \textbf{VGG-16} won the 2-nd place in ILSVRC 2014 \cite{Russakovsky2014}.

We evaluate the ``top-5 error'' using single-view testing. The view is the center $224\times224$ region cropped from the resized image whose shorter side is 256. The single-view error rate of SPP-10 is 12.51\% on the ImageNet validation set, and VGG-16 is 10.09\% in our testing (which is consistent with the number reported by \cite{Simonyan2015}\footnote{\url{http://www.vlfeat.org/matconvnet/pretrained/}}).
These numbers serve as the references for the increased error rates of our approximated models.

\begin{figure*}[t]
\begin{center}
\includegraphics[width=0.98\linewidth]{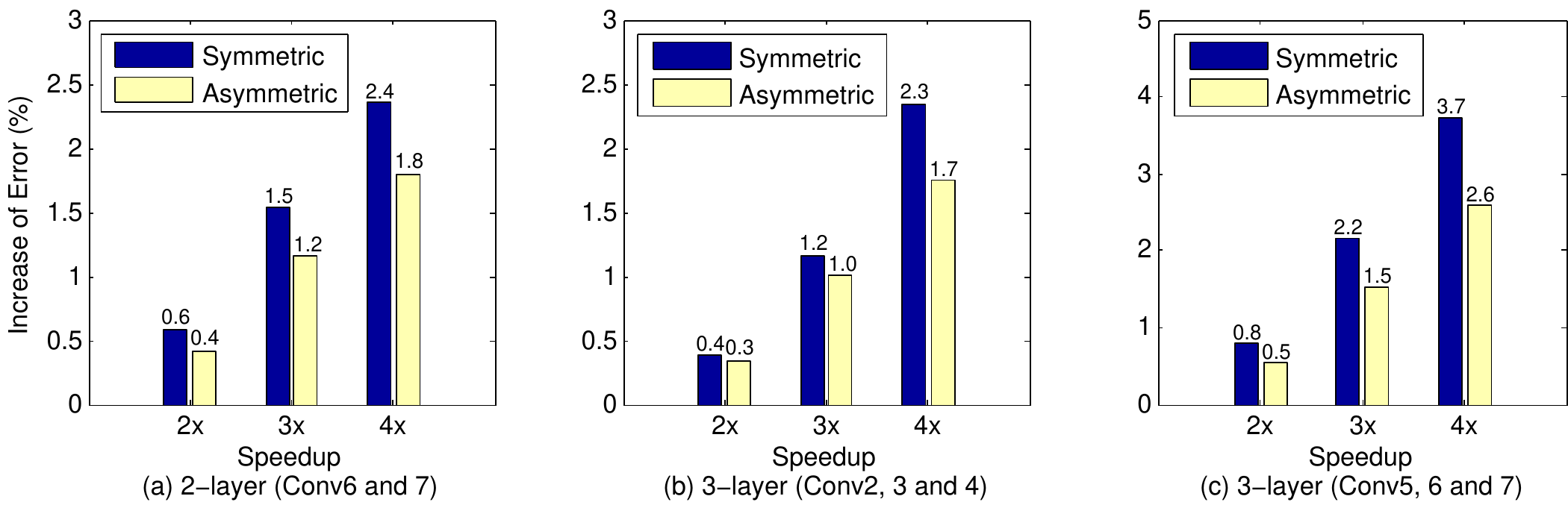}
\end{center}
   \caption{\textbf{Symmetric vs. Asymmetric} for SPP-10: the cases of 2-layer and 3-layer approximation. The speedup is computed by the complexity of the layers approximated. (a) Approximation of Conv6 \& 7. (b) Approximation of Conv2, 3 \& 4. (c) Approximation of Conv5, 6 \& 7.}
\label{fig:asym}
\end{figure*}

\subsection{Experiments with SPP-10}


We first evaluate the effect of our each step on the SPP-10 model by a series of controlled experiments. Unless specified, we do not use the 3-d decomposition.

\vspace{8pt}
\noindent\textbf{Single-Layer: Linear \vs Nonlinear}

In this subsection we evaluate the single-layer performance. When evaluating a single approximated layer, the remaining layers are unchanged and not approximated. The speedup ratio (involving that single layer only) is shown as the theoretical ratio computed by the complexity.

In Fig.~\ref{fig:layerwiseresult} we compare the performance of our linear solution (\ref{eq:pca}) and nonlinear solution (\ref{eq:relu1}). The performance is displayed as \emph{increase of error rates} (decrease of accuracy) \vs the speedup ratio of that layer. Fig.~\ref{fig:layerwiseresult} shows that the nonlinear solution consistently performs better than the linear solution. In Table~\ref{tbl:arch}, we show the sparsity (the portion of zero activations after ReLU) of each layer. A zero activation is due to the truncation of ReLU. The sparsity is over 60\% for Conv2-7, indicating that the ReLU takes effect on a substantial portion of activations. This explains the discrepancy between the linear and nonlinear solutions. Especially, the Conv7 layer has a sparsity of 95\%, so the advantage of the nonlinear solution is more obvious.

Fig.~\ref{fig:layerwiseresult} also shows that when accelerating only a single layer by 2$\times$, the increased error rates of our solutions are rather marginal or negligible. For the Conv2 layer, the error rate is increased by $<0.1\%$; for the Conv3-7 layers, the error rate is increased by $\approx0.2\%$.

We also notice that for Conv1, the degradation is negligible near $2\times$ speedup ($1.8\times$ corresponds to $d'=32$). This can be explained by Fig.~\ref{fig:pca_energy}(a): the PCA energy has little loss when $d'\geq32$.
But the degradation can grow quickly for larger speedup ratios, because in this layer the channel number $c=3$ is small and $d'$ needs to be reduced drastically to achieve the speedup ratio. So in the following whole-model experiments of SPP-10, we will use $d'=32$ for Conv1.

\renewcommand{\arraystretch}{1.35}
\begin{table*}[t]
\begin{center}
\small
\begin{tabular}{|c|c|ccccccc|c|}
\hline
speedup & rank sel. & Conv1 & Conv2 & Conv3 & Conv4 & Conv5 & Conv6 & Conv7 & err. $\uparrow \%$\\
\hline\hline
2$\times$ & no & 32 & 110 & 199 & 219 & 219 & 219 & 219 & 1.18 \\
2$\times$ & \textbf{yes} & 32 & 83 & 182 & 211 & 239 & 237 & 253 & \textbf{0.93} \\
\hline
2.4$\times$ & no & 32 & 96 & 174 & 191 & 191 & 191 & 191 & 1.77 \\
2.4$\times$ & \textbf{yes} & 32 & 74 & 162 & 187 & 207 & 205 & 219 & \textbf{1.35} \\
\hline
3$\times$ & no & 32 & 77 & 139 & 153 & 153 & 153 & 153 & 2.56 \\
3$\times$ & \textbf{yes} & 32 & 62 & 138 & 149 & 166 & 162 & 167 & \textbf{2.34} \\
\hline
4$\times$ & no & 32 & 57 & 104 & 115 & 115 & 115 & 115 & 4.32 \\
4$\times$ & \textbf{yes} & 32 & 50 & 112 & 114 & 122 & 117 & 119 & \textbf{4.20} \\
\hline
5$\times$ & no & 32 & 46 & 83 & 92 & 92 & 92 & 92 & 6.53 \\
5$\times$ & \textbf{yes} & 32 & 41 & 94 & 93 & 98 & 92 & 90 & \textbf{6.47} \\
\hline
\end{tabular}
\end{center}
\caption{\textbf{Whole-model acceleration with/without rank selection} for SPP-10. The solver is the asymmetric version. The speedup ratios shown here involve all convolutional layers (Conv1-Conv7). We fix $d'=32$ in Conv1.
In the case of no rank selection, the speedup ratio of each other layer is the same. Each column of Conv1-7 shows the rank $d'$ used, which is the number of filters after approximation. The error rates are top-5 single-view, and shown as the increase of error rates compared with no approximation.}
\label{tbl:entire_speedup}
\end{table*}

\vspace{8pt}
\noindent\textbf{Multi-Layer: Symmetric \vs Asymmetric}


Next we evaluate the performance of asymmetric reconstruction as in the problem (\ref{eq:reluasy}). We demonstrate approximating 2 layers or 3 layers. In the case of 2 layers, we show the results of approximating Conv6 and 7; and in the case of 3 layers, we show the results of approximating Conv5-7 or Conv2-4. The comparisons are consistently observed for other cases of multi-layer.

We sequentially approximate the layers involved, from a shallower one to a deeper one. In the asymmetric version (\ref{eq:reluasy}), $\ve{\hat{x}}$ is from the output of the previous approximated layer (if any), and $\ve{x}$ is from the output of the previous non-approximate layer. In the symmetric version (\ref{eq:relu}), we use $\ve{x}$ for both terms.
We have also tried another symmetric version of using $\ve{\hat{x}}$ for both terms, and found this symmetric version is even worse.

Fig.~\ref{fig:asym} shows the comparisons between the symmetric and asymmetric versions. The asymmetric solution has significant improvement over the symmetric solution. For example, when only 3 layers are approximated simultaneously (like Fig.~\ref{fig:asym} (c)), the improvement is over 1.0\% when the speedup is 4$\times$.
This indicates that the accumulative error rate due to multi-layer approximation can be effectively reduced by the asymmetric version.

When more and all layers are approximated simultaneously (as below), if without the asymmetric solution, the error rates will increase more drastically.

\vspace{8pt}
\noindent\textbf{Whole-Model: with/without Rank Selection}


In Table~\ref{tbl:entire_speedup} we show the results of whole-model acceleration. The solver is the asymmetric version.
For Conv1, we fix $d'=32$. For other layers,
when the rank selection is not used, we adopt the same speedup ratio on each layer and determine its desired rank $d'$ accordingly. When the rank selection is used, we apply it to select $d'$ for Conv2-7.
Table~\ref{tbl:entire_speedup} shows that the rank selection consistently outperforms the counterpart without rank selection. The advantage of rank selection is observed in both linear and nonlinear solutions.

In Table~\ref{tbl:entire_speedup} we notice that the rank selection often chooses a higher rank $d'$ (than the no rank selection) in Conv5-7. For example, when the speedup is 3$\times$, the rank selection assigns $d'=167$ to Conv7, while this layer only requires $d'=153$ to achieve 3$\times$ single-layer speedup of itself. This can be explained by Fig.~\ref{fig:pca_energy}(c). The energy of Conv5-7 is less concentrated, so these layers require higher ranks to achieve good approximations.

As we will show, the rank selection is more prominent for VGG-16 because of its diversity of layers.

\vspace{8pt}
\noindent\textbf{Comparisons with Jaderberg \etal's method \cite{Jaderberg2014}}

We compare with Jaderberg \etal's method \cite{Jaderberg2014}, which is a recent state-of-the-art solution to efficient evaluation. Although our \emph{decomposition} shares some high-level motivations as \cite{Jaderberg2014}, we point out that our \emph{optimization} strategy is different with \cite{Jaderberg2014} and is important for accuracy, especially for \emph{very deep} models that previous acceleration methods rarely addressed.

\begin{figure}[pb]
\begin{center}
\includegraphics[width=0.98\linewidth]{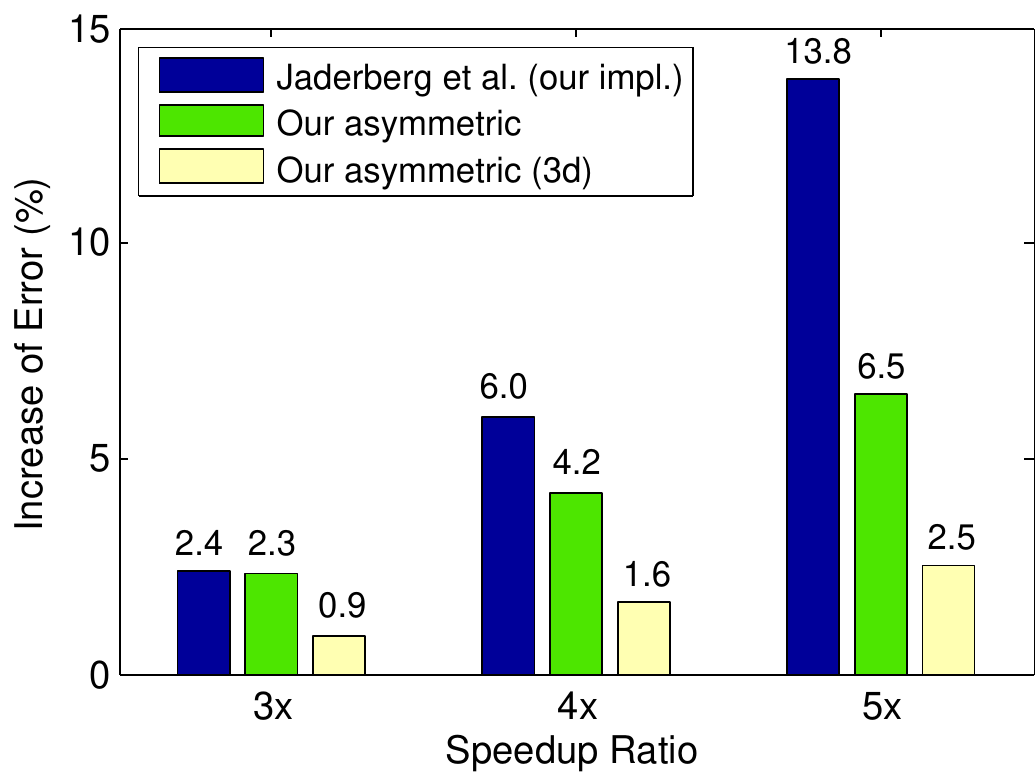}
\end{center}
   \caption{Comparisons with Jaderberg \etal's spatial decomposition method \cite{Jaderberg2014} for SPP-10. The speedup ratios are theoretical speedups of the whole model. The error rates are top-5 single-view, and shown as \textbf{the increase of error rates} compared with no approximation (\emph{smaller is better}).}
\label{fig:bars}
\end{figure}

Jaderberg \etal's method \cite{Jaderberg2014} decomposes a $k\times k$ spatial support into a cascade of $k\times 1$ and $1\times k$ spatial supports. A channel-dimension reduction is also considered.
Their optimization method focuses on the linear reconstruction error. In the paper of \cite{Jaderberg2014}, their method is only evaluated on a single layer of an OverFeat network \cite{Sermanet2014} for ImageNet.

Our comparisons are based on our implementation of \cite{Jaderberg2014}. We use the \emph{Scheme 2} decomposition in \cite{Jaderberg2014} and its ``filter reconstruction'' version (as we explain below), which is used for ImageNet as in \cite{Jaderberg2014}.
Our reproduction of the filter reconstruction in \cite{Jaderberg2014} gives a 2$\times$ single-layer speedup on Conv2 of SPP-10 with $0.2\%$ increase of error.
As a reference, in \cite{Jaderberg2014} it reports $0.5\%$ increase of error on Conv2 under a 2$\times$ single-layer speedup, evaluated on another OverFeat network \cite{Sermanet2014} similar to SPP-10.

\setlength{\tabcolsep}{8pt}
\renewcommand{\arraystretch}{1.3}
\begin{table*}[t]
\begin{center}
\begin{small}
\begin{tabular}{|c|c|c|c|c|c|}
\hline
model & \footnotesize{\tabincell{c}{ speedup \\ solution}} & \footnotesize{\tabincell{c}{top-5 err.\\(1-view)}} & \footnotesize{\tabincell{c}{CPU\\(ms)}} &
\footnotesize{\tabincell{c}{GPU\\(ms)}}
 \\
\hline
SPP-10 \cite{He2014} & - & 12.5 & 930 & 7.67 \\
\hline
\multirow{4}{*}{SPP-10 (4$\times$)}  & Jaderberg \etal \cite{Jaderberg2014} (our impl.) & 18.5 & 278 (3.3$\times$) & 2.41 (3.2$\times$) \\
& our asym. & 16.7 & 271 (3.4$\times$) & 2.62 (2.9$\times$)  \\
& our asym. (3d) & 14.1 & 267 (3.5$\times$) & 2.32 (3.3$\times$) \\
& our asym. (3d) FT & \textbf{13.8} & 267 (3.5$\times$) & 2.32 (3.3$\times$) \\
\hline
AlexNet \cite{Krizhevsky2012} & - & 18.8 & 273 & 2.37 \\
\hline
\end{tabular}
\end{small}
\end{center}
\caption{Comparisons of absolute performance of SPP-10. The top-5 error is the absolute value. The running time is a single view on a CPU (single thread, with SSE) or a GPU. The accelerated models are those of 4$\times$ theoretical speedup (Fig.~\ref{fig:bars}). On the brackets are the actual speedup ratios.}
\label{tbl:alexnet}
\end{table*}

It is worth discussing our implementation of Jaderberg \etal's \cite{Jaderberg2014} ``data reconstruction'' scheme, which was suggested to use SGD and backpropagation for optimization. In our reproduction, we find that data reconstruction works well for the \emph{character classification} task as studied in \cite{Jaderberg2014}. However, we find it nontrivial to make data reconstruction work for large models trained for ImageNet. We observe that the learning rate needs to be carefully chosen for the SGD-based data reconstruction to converge (as also reported independently in \cite{Lebedev2015} for another decomposition), and when the training starts to converge, the results are still sensitive to the initialization (for which we have tried Gaussian distributions of a wide range of variances).
We conjecture that this is because the ImageNet dataset and models are more complicated, and using SGD to regress a single layer may be sensitive to multiple local optima. In fact, Jaderberg \etal's \cite{Jaderberg2014} only report ``filter reconstruction'' results of a single layer on ImageNet. For these reasons, our implementation of Jaderberg \etal's method on ImageNet models is based on filter reconstruction. We believe that these issues have not be settled and need to be investigated further, and accelerating deep networks does not just involve \emph{decomposition} but also the way of \emph{optimization}.

In Fig.~\ref{fig:bars} we compare our method with Jaderberg \etal's \cite{Jaderberg2014} for whole-model speedup.
For whole-model speedup of \cite{Jaderberg2014}, we implement their method sequentially on Conv2-7 using the same speedup ratio.\footnote{We do not apply Jaderberg \etal's method \cite{Jaderberg2014} on Conv1, because this layer has a small number of input channels (3), and the first $k\times 1$ decomposed layer can only have a very small number of filters (\eg, 5) to approach a speedup ratio (\eg, 4$\times$). Also note that the speedup ratio is about all conv layers, and because Conv1 is not accelerated, other layers will have a slightly larger speedup.}
The speedup ratios are the theoretical complexity ratios involving all convolutional layers. Our method is the asymmetric version and with rank selection. Fig.~\ref{fig:bars} shows that when the speedup ratios are large (4$\times$ and 5$\times$), our method outperforms Jaderberg \etal's method significantly. For example, when the speedup ratio is 4$\times$, the increased error rate of our method is 4.2\%, while Jaderberg \etal's is 6.0\%. Jaderberg \etal's result degrades quickly when the speedup ratio is getting large, while ours degrades slowly. This suggests the effects of our method for reducing accumulative error.

We further compare with our asymmetric version using 3d decomposition (Sec.~\ref{sec:hd}). In Fig.~\ref{fig:bars} we show the results ``\emph{asymmetric (3d)}''.
Fig.~\ref{fig:bars} shows that
this strategy leads to significantly smaller increase of error. For example, when the speedup is 5$\times$, the error is increased by only 2.5\%.
Our asymmetric solver effectively controls the accumulative error even if the multiple layers are decomposed extensively, and the 3d decomposition is easier to achieve a certain speedup ratio.

For completeness, we also evaluate our approximation method on the \emph{character classification} model released by \cite{Jaderberg2014}. Our asymmetric (3d) solution achieves 4.5$\times$ speedup with only a drop of 0.7\% in classification accuracy, which is better than the 1\% drop for the same speedup reported by \cite{Jaderberg2014}.

\vspace{8pt}
\noindent\textbf{Comparisons with Training from Scratch}

The architecture of the approximated model can also be trained ``\emph{from scratch}'' on the ImageNet dataset. One hypothesis is that the underlying architecture is sufficiently powerful, and the acceleration algorithm might be not necessary. We show that this hypothesis is premature.

We directly train the model of the same architecture as the decomposed model. The decomposed model is much deeper than the original model (each layer replaced by three layers), so we adopt the initialization method in \cite{He2015} otherwise it is not easy to converge. We train the model for 100 epochs. We follow the common practice in \cite{Chatfield2014,He2014} of training ImageNet models.

The comparisons are in Table~\ref{tbl:scratch}. The accuracy of the model trained from scratch is worse than that of our accelerated model by a considerable margin (2.8\%). These results indicate that the accelerating algorithms can effectively digest information from the trained models. They also suggest that the models trained from scratch have much redundancy.

\setlength{\tabcolsep}{6pt}
\renewcommand{\arraystretch}{1.3}
\begin{table}[h]
\begin{center}
\begin{small}
\begin{tabular}{|c|c|c|}
\hline
model & \footnotesize{\tabincell{c}{top-5 err.\\(1-view)}} & \footnotesize{\tabincell{c}{increased err.\\(1-view)}}  \\
\hline
SPP-10 \cite{He2014} & 12.5 & - \\
\hline
our asym. 3d (4$\times$) & 14.1 & 1.6 \\
from scratch & 16.9 & 4.4 \\
\hline
\end{tabular}
\end{small}
\end{center}
\caption{Comparisons with the same decomposed architecture trained from scratch. }
\label{tbl:scratch}
\end{table}

\renewcommand{\arraystretch}{1.3}
\begin{table*}[t]
\begin{center}
\small
\begin{tabular}{|c|c|c|c|c|c|c|c|c|}
\hline
layer & filter size & \# channels & \# filters & stride & output size & complexity (\%) & \# of zeros \\
\hline
Conv1$_1$ & 3 $\times$ 3 & 3 & 64 & 1 & 224 $\times$ 224 & 0.6 & 0.48 \\
Conv1$_2$ & 3 $\times$ 3 & 64 & 64 & 1 & 224 $\times$ 224 & 12.0 & 0.32 \\
Pool1 & 3 $\times$ 3 &   &    & 2 & 112 $\times$ 112   &     &      \\
\hline
Conv2$_1$ & 3 $\times$ 3 & 64 & 128 & 1 & 112 $\times$ 112 & 6.0 & 0.35 \\
Conv2$_2$ & 3 $\times$ 3 & 128 & 128 & 1 & 112 $\times$ 112 & 12.0 & 0.52 \\
Pool2 & 2 $\times$ 2 &   &    & 2 & 56 $\times$ 56   &     &      \\
\hline
Conv3$_1$ & 3 $\times$ 3 & 128 & 256 & 1 & 56 $\times$ 56 & 6.0 & 0.48 \\
Conv3$_2$ & 3 $\times$ 3 & 256 & 256 & 1 & 56 $\times$ 56 & 12.1 & 0.48 \\
Conv3$_3$ & 3 $\times$ 3 & 256 & 256 & 1 & 56 $\times$ 56 & 12.1 & 0.70 \\
Pool3 & 2 $\times$ 2 &   &    & 2 & 28 $\times$ 28   &     &      \\
\hline
Conv4$_1$ & 3 $\times$ 3 & 256 & 512 & 1 & 28 $\times$ 28 & 6.0 & 0.65 \\
Conv4$_2$ & 3 $\times$ 3 & 512 & 512 & 1 & 28 $\times$ 28 & 12.1 & 0.70 \\
Conv4$_3$ & 3 $\times$ 3 & 512 & 512 & 1 & 28 $\times$ 28 & 12.1 & 0.87 \\
Pool4 & 2 $\times$ 2 &   &    & 2 & 14 $\times$ 14   &     &      \\
\hline
Conv5$_1$ & 3 $\times$ 3 & 512 & 512 & 1 & 14 $\times$ 14 & 3.0 & 0.76 \\
Conv5$_2$ & 3 $\times$ 3 & 512 & 512 & 1 & 14 $\times$ 14 & 3.0 & 0.80 \\
Conv5$_3$ & 3 $\times$ 3 & 512 & 512 & 1 & 14 $\times$ 14 & 3.0 & 0.93 \\
\hline

\end{tabular}
\end{center}
\caption{The architecture of the VGG-16 model \cite{Simonyan2015}. It has 13 conv layers and 3 fc layers. The column ``complexity'' is the theoretical time complexity, shown as relative numbers to the total convolutional complexity. The column ``\# of zeros'' is the relative portion of zero responses, which shows the ``sparsity'' of the layer.}
\label{tbl:arch_vgg16}
\end{table*}

\setlength{\tabcolsep}{6pt}
\renewcommand{\arraystretch}{1.3}
\begin{table*}[t]
\begin{center}
\small
\begin{tabular}{|c|c|cc|cc|ccc|ccc|ccc|c|}
\hline
speedup & rank sel. & C1$_1$ & C1$_2$ & C2$_1$ & C2$_2$ & C3$_1$ & C3$_2$ & C3$_3$ & C4$_1$ & C4$_2$ & C4$_3$ & C5$_1$ & C5$_2$ & C5$_3$ &err. $\uparrow \%$\\
\hline\hline
2$\times$ & no & 64 & 28 & 52 & 57 & 104 & 115 & 115 & 209 & 230 & 230 & 230 & 230 & 230 & 0.99 \\
2$\times$ & \textbf{yes} & 64 & 18 & 41 & 50 & 94 & 96 & 116 & 207 & 213 & 260 & 467 & 455 & 442 & \textbf{0.28} \\
\hline
3$\times$ & no & 64 & 19 & 34 & 38 & 69 & 76 & 76 & 139 & 153 & 153 & 153 & 153 & 153 & 3.25 \\
3$\times$ & \textbf{yes} & 64 & 15 & 31 & 34 & 68 & 64 & 75 & 134 & 126 & 146 & 312 & 307 & 294 & \textbf{1.66} \\
\hline
4$\times$ & no & 64 & 14 & 26 & 28 & 52 & 57 & 57 & 104 & 115 & 115 & 115 & 115 & 115 & 6.38 \\
4$\times$ & \textbf{yes} & 64 & 11 & 25 & 28 & 52 & 46 & 56 & 104 & 92 & 100 & 232 & 224 & 214 & \textbf{3.84} \\
\hline
\end{tabular}
\end{center}
\caption{\textbf{Whole-model acceleration with/without rank selection} for VGG-16. The solver is the asymmetric version. The speedup ratios shown here involve all convolutional layers.  We do not accelerate Conv1$_1$.
In the case of no rank selection, the speedup ratio of each other layer is the same. Each column of C1$_2$-C5$_3$ shows the rank $d'$ used, which is the number of filters after approximation. The error rates are top-5 single-view, and shown as the increase of error rates compared with no approximation.}
\label{tbl:entire_speedup_vgg16}
\end{table*}

\vspace{8pt}
\noindent\textbf{Comparisons of Absolute Performance}

Table~\ref{tbl:alexnet} shows the comparisons of the absolute performance of the accelerated models. We also evaluate the AlexNet \cite{Krizhevsky2012} which is similarly fast as our accelerated 4$\times$ models.
The comparison is based on our re-implementation of AlexNet. Our AlexNet is the same as in \cite{Krizhevsky2012} except that the GPU splitting is ignored.
Our re-implementation of this model has top-5 single-view error rate as 18.8\% (10-view top-5 16.0\% and top-1 37.6\%). This is better than the one reported in \cite{Krizhevsky2012}\footnote{In \cite{Krizhevsky2012} the 10-view error is top-5 18.2\% and top-1 40.7\%.}.

The models accelerated by our asymmetric (3d) version have 14.1\% and 13.8\% top-5 error, without and with fine-tuning. This means that the accelerated model has 5.0\% lower error than AlexNet, while its speed is nearly the same as AlexNet.

Table~\ref{tbl:alexnet} also shows the actual running time per view, on a C++ implementation and Intel i7 CPU (2.9GHz) or Nvidia K40 GPU.
In our CPU version, our method has actual speedup ratios (3.5$\times$) close to theoretical speedup ratios (4.0$\times$). This overhead mainly comes from the fc and other layers. In our GPU version, the actual speedup ratio is about 3.3$\times$. An accelerated model is less easy for parallelism in a GPU, so the actual ratio is lower.

\subsection{Experiments with VGG-16}

The very deep VGG models \cite{Simonyan2015} have substantially improved a wide range of visual recognition tasks, including object detection \cite{Girshick2014,Girshick2015,Ren2015,Ren2015b}, semantic segmentation \cite{Long2015,Dai2015,Hariharan2015,Chen2015a,Dai2015a}, image captioning \cite{Fang2015,Karpathy2015,Chen2015}, video/action recognition \cite{Srivastava2015}, image question answering \cite{Ren2015a}, texture recognition \cite{Cimpoi2015}, \etc. Considering the big impact yet slow speed of this model, we believe it is of practical significance to accelerate this model.

\setlength{\tabcolsep}{12pt}
\renewcommand{\arraystretch}{1.3}
\begin{table*}[t]
\begin{center}
\begin{small}
\begin{tabular}{|c|c|c|c|}
\hline
 \multicolumn{4}{|c|}{increase of top-5 error (1-view)}\\
\hline
speedup ratio & 3$\times$ & 4$\times$ & 5$\times$\\
\hline
Jaderberg \etal \cite{Jaderberg2014} (our impl.) & 2.3 & 9.7 & 29.7 \\
our asym. (3d) & 0.4 & 0.9 & 2.0 \\
our asym. (3d) FT & \textbf{0.0} & \textbf{0.3} & \textbf{1.0} \\
\hline
\end{tabular}
\end{small}
\end{center}
\caption{Accelerating the \textbf{VGG-16} model \cite{Simonyan2015} using a speedup ratio of 3$\times$, 4$\times$, or 5$\times$. The top-5 error rate (1-view) of the VGG-16 model is 10.1\%. This table shows the increase of error on this baseline.}
\label{tbl:vggnet}
\end{table*}

\setlength{\tabcolsep}{8pt}
\renewcommand{\arraystretch}{1.3}
\begin{table*}[t]
\begin{center}
\begin{small}
\begin{tabular}{|c|c|c|c|c|}
\hline
model & \footnotesize{\tabincell{c}{ speedup \\ solution}} & \footnotesize{\tabincell{c}{top-5 error\\(1-view)}} & \footnotesize{\tabincell{c}{CPU\\(ms)}} &
\footnotesize{\tabincell{c}{GPU\\(ms)}}
 \\
\hline
VGG-16 \cite{Simonyan2015} & - & 10.1 & 3287 & 18.60 \\
\hline
\multirow{4}{*}{VGG-16 (4$\times$)}  & Jaderberg \etal \cite{Jaderberg2014} (our impl.) & 19.8 & 875 (3.8$\times$) & 6.40 (2.9$\times$) \\
& our asym. & 13.9 & 875 (3.8$\times$) & 7.97 (2.3$\times$) \\
& our asym. (3d) & 11.0 & 860 (3.8$\times$) & 6.30 (3.0$\times$) \\
& our asym. (3d) FT & \textbf{10.4} & 858 (3.8$\times$) & 6.39 (2.9$\times$) \\
\hline
\end{tabular}
\end{small}
\end{center}
\caption{Absolute performance of accelerating the \textbf{VGG-16} model \cite{Simonyan2015}. The top-5 error is the absolute value. The running time is a single view on a CPU (single thread, with SSE) or a GPU. The accelerated models are those of 4$\times$ theoretical speedup (Table~\ref{tbl:vggnet}). On the brackets are the actual speedup ratios.}
\label{tbl:vggnet_abs}
\end{table*}

\vspace{6pt}
\noindent\textbf{Accelerating VGG-16 for ImageNet Classification}

Firstly we discover that our whole-model \emph{rank selection} is particularly important for accelerating VGG-16. In Table~\ref{tbl:entire_speedup_vgg16} we show the results without/with rank selection. No 3d decomposition is used in this comparison. For a 4$\times$ speedup, the rank selection reduces the increased error from 6.38\% to 3.84\%. This is because of the \textbf{greater diversity of layers} in VGG-16 (Table~\ref{tbl:arch_vgg16}). Unlike SPP-10 (or other shallower models \cite{Krizhevsky2012,Zeiler2014}) that repeatedly applies 3$\times$3 filters on the same feature map size, the VGG-16 model applies them more evenly on five feature map sizes (224, 112, 56, 28, and 14). Besides, as the filter numbers in Conv5$_1$-5$_3$ are not increased, the time complexity of Conv5$_1$-5$_3$ is smaller than others.
The selected ranks $d'$ in Table~\ref{tbl:entire_speedup_vgg16} show their adaptivity - \eg, the layers Conv5$_1$ to Conv5$_3$ keep more filters, because they have small time complexity and it is not a good trade-off to compactly reduce them. The whole-model rank selection is a key to maintain a high accuracy for accelerating VGG-16.

In Table~\ref{tbl:vggnet} we evaluate our method on VGG-16 for ImageNet classification. Here we evaluate our asymmetric 3d version (without or with fine-tuning). We evaluate challenging speedup ratios of 3$\times$, 4$\times$ and 5$\times$. The ratios are those of the theoretical speedups of all 13 conv layers. 

Somewhat surprisingly, our method has demonstrated compelling results for this very deep model, even without fine-tuning. Our no-fine-tuning model has a 0.9\% increase of 1-view top-5 error for a speedup ratio of 4$\times$. On the contrary, the previous method \cite{Jaderberg2014} suffers greatly from the increased depth because of the rapidly accumulated error of multiple approximated layers. After fine-tuning, our model has a 0.3\% increase of 1-view top-5 error for a 4$\times$ speedup. This degradation is even lower than that of the shallower model of SPP-10. This suggests that the information in the very deep VGG-16 model is highly redundant, and our method is able to effectively digest it.

\begin{figure}[b]
\begin{center}
\includegraphics[width=0.7\linewidth]{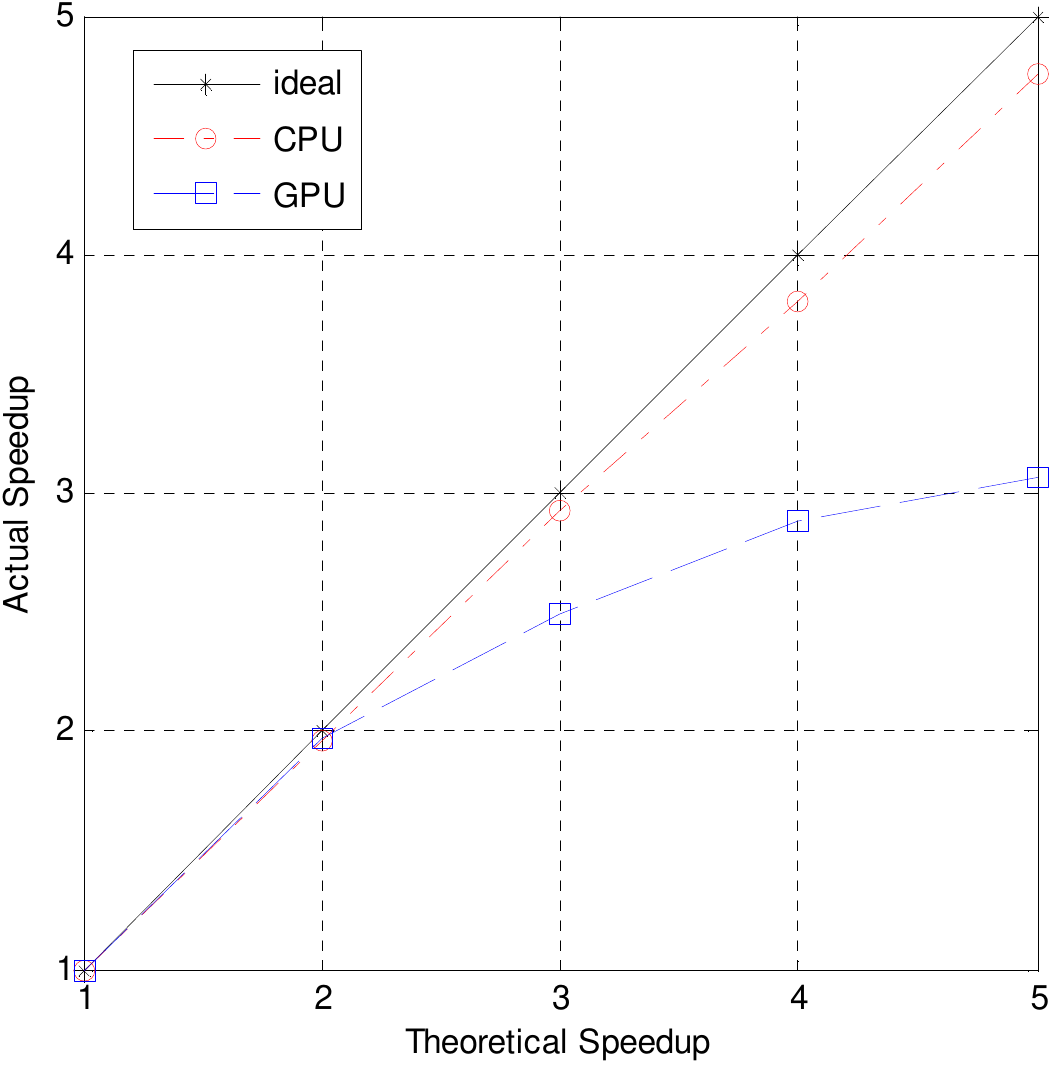}
\end{center}
   \caption{Actual \vs theoretical speedup ratios of VGG-16 using CPU and GPU implementations.}
\label{fig:speedup}
\end{figure}

Fig.~\ref{fig:speedup} shows the actual \vs theoretical speedup ratios of VGG-16 using CPU and GPU implementations. The CPU speedup ratios are very close to the theoretical ratios. The GPU implementation, which is based on the standard Caffe library \cite{Jia2014}, exhibits a gap between actual \vs theoretical ratios (as is also witnessed in \cite{Figurnov2015}). GPU speedup ratios are more sensitive to specialized implementation, and the generic Caffe kernels are not optimized for some layers (\eg, 1$\times$1, 1$\times$3, and 3$\times$1 convolutions). We believe that a more specially engineered implementation will increase the actual GPU speedup ratio.

Figurnov \etal's work \cite{Figurnov2015} is one of few existing works that present results of accelerating the whole model of VGG-16. They report increased top-5 1-view error rates of 3.4\% and 7.1\% for actual CPU speedups of 3$\times$ and 4$\times$ (for 4$\times$ theoretical speedup they report a 3.8$\times$ actual CPU speedup). Thus our method is substantially more accurate than theirs. Note that results in \cite{Figurnov2015} are after fine-tuning. This suggests that fine-tuning is not sufficient for whole-model acceleration; a good optimization solver for the decomposition is needed.

\vspace{6pt}
\noindent\textbf{Accelerating VGG-16 for Object Detection}

Current state-of-the-art object detection results \cite{Girshick2014,Girshick2015,Ren2015,Ren2015b} mostly rely on the VGG-16 model. We evaluate our accelerated VGG-16 models for object detection. Our method is based on the recent Fast R-CNN \cite{Girshick2015}.

We evaluate on the PASCAL VOC 2007 object detection benchmark \cite{Everingham2007}. This dataset contains 5k trainval images and 5k test images. We follow the default setting of Fast R-CNN using the publicly released code\footnote{\url{https://github.com/rbgirshick/fast-rcnn}}.
We train Fast R-CNN on the trainval set and evaluate on the test set. The accuracy is evaluated by mean Average Precision (mAP).

In our experiments, we first approximate the VGG-16 model on the ImageNet classification task. Then we use the approximated model as the pre-trained model for Fast R-CNN. We use our asymmetric 3d version with fine-tuning.
Note that unlike image classification where the conv layers dominate running time, for Fast R-CNN detection the conv layers consume about 70\% actual running time \cite{Girshick2015}. The reported speedup ratios are the theoretical speedups about the conv layers only.

Table~\ref{tbl:fastrcnn} shows the results of the accelerated models in PASCAL VOC 2007 detection. Our method with a 4$\times$ convolution speedup has a graceful degradation of \textbf{0.8\%} in mAP. We believe this trade-off between accuracy and speed is of practical importance, because even with the recent advance of fast object detection \cite{He2014,Girshick2015}, the feature extraction running time is still considerable.

\setlength{\tabcolsep}{6pt}
\renewcommand{\arraystretch}{1.3}
\begin{table}[t]
\begin{center}
\begin{small}
\begin{tabular}{|c|c|c|}
\hline
conv speedup & mAP & $\Delta$mAP \\
\hline
baseline & 66.9 & - \\
\hline
3$\times$ & 66.9 & 0.0 \\
4$\times$ & 66.1 & -0.8 \\
5$\times$ & 65.2 & -1.7 \\
\hline
\end{tabular}
\end{small}
\end{center}
\caption{\textbf{Object detection mAP} on the PASCAL VOC 2007 test set. The detector is Fast R-CNN \cite{Girshick2015} using the pre-trained VGG-16 model. }
\label{tbl:fastrcnn}
\end{table}

\section{Conclusion}

We have presented an acceleration method for very deep networks. Our method is evaluated under whole-model speedup ratios. It can effectively reduce the accumulated error of multiple layers thanks to the nonlinear asymmetric reconstruction.
Competitive speedups and accuracy are demonstrated in the complex ImageNet classification task and PASCAL VOC object detection task.

\bibliographystyle{IEEEtran}
\bibliography{IEEEabrv,cnn_speedup}
\end{document}